# Fuzzy Soft Set Theory based Expert System for the Risk Assessment in Breast Cancer Patients

Liaqat Muhammad Sheharyar

Nanjing University of Science & Technology

## Abstract

Breast cancer remains one of the leading causes of mortality among women worldwide, with early diagnosis being critical for effective treatment and improved survival rates. However, timely detection continues to be a challenge due to the complex nature of the disease and variability in patient risk factors. This study presents a fuzzy soft set theory–based expert system designed to assess the risk of breast cancer in patients using measurable clinical and physiological parameters. The proposed system integrates Body Mass Index (BMI), Insulin Level (IL), Leptin Level (LL), Adiponectin Level (AL), and age as input variables to estimate breast cancer risk through a set of fuzzy inference rules and soft set computations. These parameters can be obtained from routine blood analyses, enabling a non-invasive and accessible method for preliminary assessment. The dataset used for model development and validation was obtained from the UCI Machine Learning Repository. The proposed expert system aims to support healthcare professionals in identifying high-risk patients and determining the necessity of further diagnostic procedures such as biopsies.

## 1. INTRODUCTION

Cancer is a term that refers to uncontrolled cell growth in a specific area of the body. Tumors are lumps formed by abnormal cell growth. Breast cancer is the most prevalent type of invasive cancer in women and the second leading cause of death from cancer in women, behind lung cancer [1]. To address this issue effectively, advanced computer technology should be used. Using new technology, medical science is on the lookout for solutions that will benefit the medical community in every way [2]. To address these medical queries, expert systems, Case Based Reasoning, data mining, and other machine learning techniques are used [3-5]. Because uncertainty exists during diagnosis, fuzzy set and soft set theory are critical for dealing with vagueness [6-10].

The possible advantages of an Expert System capable of making timely and accurate skin disease diagnoses are enticing. Expert systems have been used in medical diagnosis for decades. An expert system is a computer program that takes disease symptoms and demographic information about patients as input and outputs information about diseases. The primary objective of this research is to develop a soft expert system for breast cancer diagnosis. In the field of medical science, implementations based on fuzzy logic improve the detection capability of various human diseases [11-13]. Traditionally, expert systems for diagnosis use bivalent logic, which means that symptoms can only be entered in a Yes/No format. However, by combining fuzzy logics and soft sets with expert systems, the smallest fuzzy soft set to answer ranges between 1 and 0 is calculated using mathematical reasoning [5, 14, 15]. A fuzzy diagnosis system captures both the symptoms and severity of a disease. To ensure an accurate and efficient diagnosis, symptoms should be classified according to their degree of truthiness or severity, e.g., "Mild," "Low," "High," and "Very high." As a result, we develop an expert system using fuzzy logic. The dataset for this study was obtained from the University of California, Irvine's Machine Learning Repository [16]. We use five attributes and the data of ten randomly selected patients. To calibrate this proposed concept, we will use the fuzzy inference system (FIS) tool in the MATLAB software.

The remainder of this study is organized in the following manner. Section 2 presents material and methods for identifying the advantages and disadvantages of using fuzzy inference systems for breast cancer detection. Later in section 3, the fuzzification process is described in detail. Section 4 describes how fuzzy sets are transformed into soft sets that accept five standard input parameters (Age, Body Mass Index, Insulin, Leptin, Adiponectin). Section 5 will discuss normal parameter reduction for fuzzy soft sets. Section 6 will demonstrate how Kong's algorithm can be used to forecast which patients are likely to develop breast cancer. Sections 7 and 8 will present the comparison table and score table in which we will examine our observations and their implications. The paper concludes with a results and discussion section that demonstrates the paper's most significant findings, limitations, and directions for future research.

## 2. MATERIAL AND METHODS

The Breast cancer data set obtained from the UCI Machine Learning Repository contained 116 instances initially. We randomly select ten instances to evaluate the system's performance. Body Mass Index (BMI), Insulin Level (IL), Leptin Level (LL), Adiponectin Level (AL), and age are used as input or independent variables. Our system generates a degree risk factor for breast cancer.

The steps involved in developing our designed system are depicted in Figure 1.

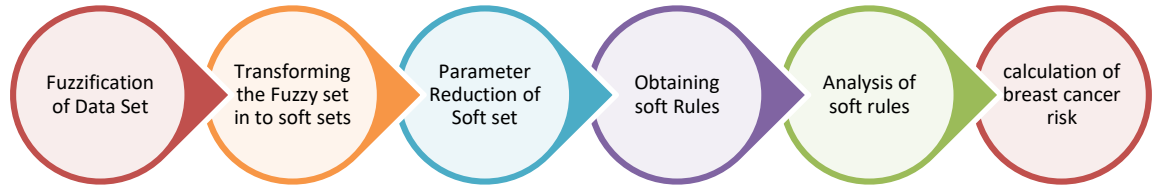

*Figure 1: Steps of Soft Expert System*

## 3. FIRST STEP: FUZZIFICATION OF DATA SET

Because the data set is inconvenient for applying directly to soft sets, we first fuzzify it. To fuzzify the factors, we used the input values from ten randomly chosen medical record, as shown in Table 1.

*TABLE 1: SAMPLE DATA OF 10 PATIENTS*

| Sample No | Age | BMI | Insulin | Leptin | Adiponectin |
|---|---|---|---|---|---|
| $\mu_3$ | 82 | 23.12 | 4.50 | 17.94 | 22.43 |
| $\mu_{11}$ | 49 | 23.01 | 5.66 | 35.59 | 26.72 |
| $\mu_{19}$ | 64 | 34.53 | 4.43 | 21.21 | 5.46 |
| $\mu_{31}$ | 66 | 36.21 | 15.53 | 74.71 | 7.54 |
| $\mu_{45}$ | 71 | 30.30 | 8.34 | 56.50 | 8.13 |
| $\mu_{60}$ | 62 | 22.66 | 3.48 | 9.86 | 11.24 |
| $\mu_{71}$ | 44 | 24.74 | 58.46 | 18.16 | 16.10 |
| $\mu_{82}$ | 71 | 25.51 | 10.40 | 19.07 | 5.49 |
| $\mu_{91}$ | 82 | 31.22 | 18.08 | 31.65 | 9.92 |
| $\mu_{104}$ | 57 | 34.84 | 12.55 | 33.16 | 2.36 |

### 3.1 INPUT VARIABLES

These are variables that are comparable to the disease risk. We must first fuzzify these variables before we can use them in a soft system. The process of variable fuzzification is detailed below.

#### A. AGE

This input field is partitioned into four fuzzy sets (Child, Young, Mild and Old). Table 2 will list these fuzzy sets and their associated ranges. The membership functions of "Child" and "old" are trapezoidal, while those of "Young" and "Mild" are triangular. Figure 2 illustrates the membership functions, while Eq. illustrates the membership expressions (1). The age variable is shown in table 3 as a fuzzy soft set.

*Table 2: Classification of Age*

| Input Field | Range | Fuzzy Sets |
|---|---|---|
| Age | 1-15 | Child |
| | 10-40 | Young |
| | 30-60 | Mild |
| | 55-70 | Old |

##### a. MEMBERSHIP FUNCTIONS OF AGE

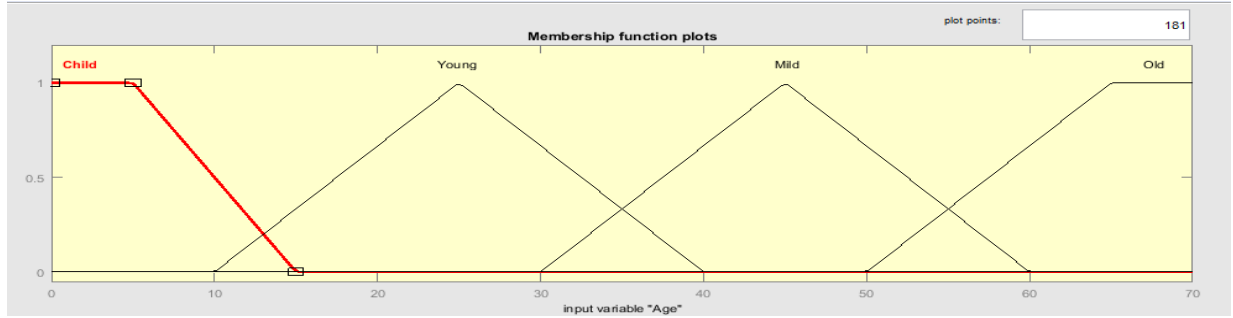

*Figure 2: Membership Functions of Age*

### b. FUZZY MEMBERSHIP EXPRESSIONS (1) FOR AGE

$$\mu_{Child}(x) = \begin{cases} 1 & x < 5 \\ \frac{15-x}{10} & 5 \leq x < 15 \end{cases}$$

$$\mu_{Young}(x) = \begin{cases} \frac{x-10}{15} & 10 \leq x < 25 \\ 1 & x = 25 \\ \frac{40-x}{15} & 25 \leq x < 40 \end{cases}$$

$$\mu_{Mild}(x) = \begin{cases} \frac{x-30}{15} & 30 \leq x < 45 \\ 1 & x = 45 \\ \frac{60-x}{15} & 45 \leq x < 60 \end{cases}$$

$$\mu_{Old}(x) = \begin{cases} \frac{x-50}{15} & 50 \leq x < 65 \\ 1 & x \geq 65 \end{cases}$$

### c. FUZZY SOFT SET OF AGE ($\tilde{L}, A$)

Table 3 illustrates a fuzzy soft set for the age variable.

*Table 3: Fuzzy Soft Set of Age ($\tilde{L}, A$)*

| Sample No | $(AGE)_C = \alpha_1$ | $(AGE)_Y = \alpha_2$ | $(AGE)_C = \alpha_3$ | $(AGE)_C = \alpha_4$ |
|---|---|---|---|---|
| $\mu_3$ | 0.00 | 0.00 | 0.00 | 1.00 |
| $\mu_{11}$ | 0.00 | 0.00 | 0.73 | 0.00 |
| $\mu_{19}$ | 0.00 | 0.00 | 0.00 | 0.93 |
| $\mu_{31}$ | 0.00 | 0.00 | 0.00 | 1.00 |
| $\mu_{45}$ | 0.00 | 0.00 | 0.00 | 1.00 |
| $\mu_{60}$ | 0.00 | 0.00 | 0.00 | 0.80 |
| $\mu_{71}$ | 0.00 | 0.00 | 0.93 | 0.00 |
| $\mu_{82}$ | 0.00 | 0.00 | 0.00 | 1.00 |
| $\mu_{91}$ | 0.00 | 0.00 | 0.00 | 1.00 |
| $\mu_{104}$ | 0.00 | 0.00 | 0.20 | 0.46 |

### B. BODY MASS INDEX

BMI is an abbreviation for body mass index. individuals over the age of 25 have an increased risk of being diagnosed with breast cancer. Obesity classes I (OI), II (OII), and III (OIII) have been classified as three fuzzy sets, as shown in table 4. For obesity classes I (OI) and III (OIII), triangular membership functions are used, while for obesity class II (OII), triangular membership functions are used. In Equations, membership expressions are defined (2). The following table illustrates a fuzzy soft set of the BMI variable.

*Table 4: Classification of Body Mass Index*

| Symptom | Range | Fuzzy Division |
|---|---|---|
| | 0-22 | Obesity Class I (OI), |
| Body Mass Index | 20-33 | Obesity Class II (OII), |
| | 30-40 | Obesity Class III (OIII), |

### a. MEMBERSHIP FUNCTIONS OF BODY MASS INDEX

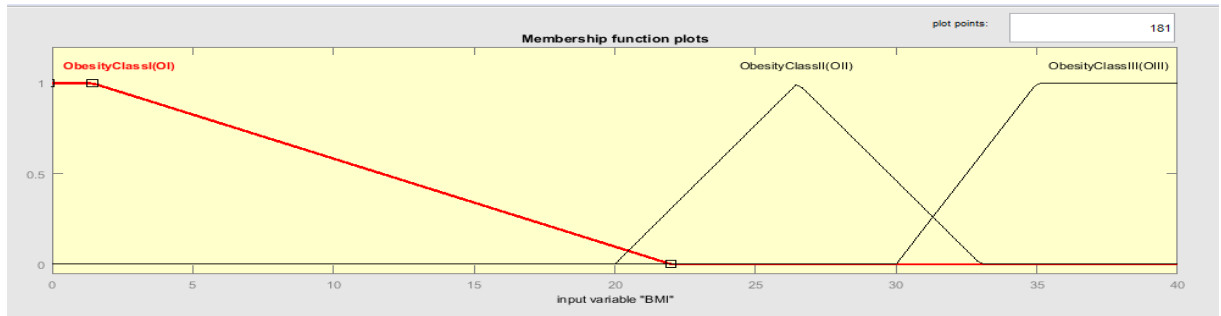

*Figure 3: Membership functions of Body Mass Index*

### b. EQUATIONS (2) OF MEMBERSHIP EXPRESSIONS FOR BODY MASS INDEX

$$\mu \text{ obesity class I (x)} = \begin{cases} 1 & x < 2 \\ \frac{22-x}{20} & 2 \leq x < 22 \end{cases}$$

$$\mu \text{ obesity class II (x)} = \begin{cases} \frac{x-20}{6} & 20 \leq x < 26 \\ 1 & x = 26 \\ \frac{33-x}{7} & 26 \leq x < 33 \end{cases}$$

$$\mu \text{ obesity class III (x)} = \begin{cases} \frac{x-30}{5} & 30 \leq x < 35 \\ 1 & x \geq 35 \end{cases}$$

### c. FUZZY SOFT SET OF BODY MASS INDEX ($\tilde{M}$,B)

Table 5 illustrates a fuzzy soft set for the BMI variable.

*Table 5: Fuzzy Soft Set of Body Mass Index ($\tilde{M}$,B)*

| Sample No | (BMI)$_{OI}$= $\beta_1$ | (BMI)$_{OII}$= $\beta_2$ | (BMI)$_{OIII}$= $\beta_3$ |
|---|---|---|---|
| $\mu_3$ | 0.00 | 0.50 | 0.00 |
| $\mu_{11}$ | 0.00 | 0.50 | 0.00 |
| $\mu_{19}$ | 0.00 | 0.00 | 0.80 |
| $\mu_{31}$ | 0.00 | 0.00 | 1.00 |
| $\mu_{45}$ | 0.00 | 0.83 | 0.00 |
| $\mu_{60}$ | 0.00 | 0.33 | 0.00 |
| $\mu_{71}$ | 0.00 | 0.80 | 0.00 |
| $\mu_{82}$ | 0.00 | 0.83 | 0.00 |
| $\mu_{91}$ | 0.00 | 0.35 | 0.24 |
| $\mu_{104}$ | 0.00 | 0.00 | 0.96 |

### C. INSULIN

Excessive insulin and/or C-peptide (a marker of insulin secretion) levels are associated with a higher risk of relapse and mortality in women with early-stage breast cancer, even in the absence of diabetes. In table 6, hypoglycemia, normal glucose, and hyperinsulinemia have been classified as three fuzzy sets. For Hypoglycemia and Hyperinsulinemia, fuzzy sets, trapezoidal membership functions were chosen, while for Normal, a triangular membership function was chosen. In Equations, membership expressions are defined (3). The variable Insulin is represented by a fuzzy soft set in Table 7.

*Table 6: Classification of Insulin*

| Symptom | Range | Fuzzy Division |
|---|---|---|
| *Insulin* | 0-5 | Hypoglycemia |
| | 3-10 | Normal |
| | 8-40 | Hyperinsulinemia |

### a. MEMBERSHIP FUNCTIONS OF INSULIN

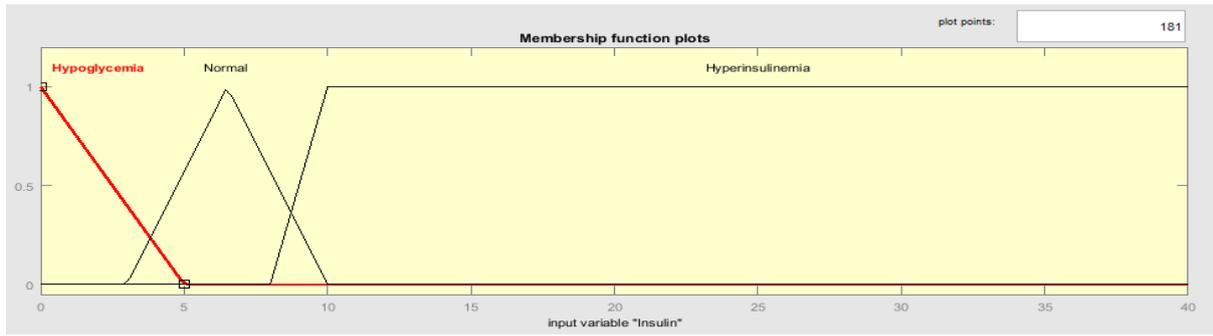

*Figure 4: Membership functions of Insulin*

### b. EQUATIONS (3) OF MEMBERSHIP EXPRESSIONS FOR INSULIN (Ñ,C)

$$\mu \text{ Hypoglycemia } (x) = \begin{cases} 1 & x < 0 \\ \frac{5-x}{5} & 0 \leq x < 5 \end{cases}$$

$$\mu \text{ Normal } (x) = \begin{cases} \frac{x-3}{3.5} & 3 \leq x < 6.5 \\ 1 & x = 6.5 \\ \frac{10-x}{3.5} & 6.5 \leq x < 10 \end{cases}$$

$$\mu \text{ Hyperinsulinemia } (x) = \begin{cases} \frac{x-8}{2} & 8 \leq x < 10 \\ 1 & x \geq 10 \end{cases}$$

### c. FUZZY SOFT SET OF INSULIN (N,C)

Table 7 illustrates a fuzzy soft set for the Insulin variable.

*Table 6: Fuzzy Soft Set of Insulin (Ñ,C)*

| Sample No | $(INS)_L = \gamma_1$ | $(INS)_M = \gamma_2$ | $(INS)_H = \gamma_3$ |
|---|---|---|---|
| $\mu_3$ | 0.10 | 0.42 | 0.00 |
| $\mu_{11}$ | 0.00 | 0.76 | 0.00 |
| $\mu_{19}$ | 0.11 | 0.40 | 0.00 |
| $\mu_{31}$ | 0.00 | 0.00 | 1.00 |
| $\mu_{45}$ | 0.00 | 0.47 | 0.14 |
| $\mu_{60}$ | 0.43 | 0.13 | 0.00 |
| $\mu_{71}$ | 0.00 | 0.00 | 1.00 |
| $\mu_{82}$ | 0.00 | 0.00 | 1.00 |
| $\mu_{91}$ | 0.00 | 0.00 | 1.00 |
| $\mu_{104}$ | 0.00 | 0.00 | 1.00 |

### D. LEPTIN

Breast cancer patients had a higher leptin level than breast benign controls. Leptin is classified into four fuzzy sets, as illustrated in Table 8. As illustrated in Fig 5, a trapezoidal membership function is used to symbolize Medium-Leptin and High-Leptin sets, while a triangular membership function is used to symbolize Low-Leptin and Very High-Leptin sets. Scaling fuzzy membership expressions were defined in Equation (4). The fuzzy soft set for leptin is shown in Table 9.

*Table 7: Classification of Leptin*

| Symptom | Range | Fuzzy Division |
|---|---|---|
| Leptin | 0-20 | Low-Leptin |
| | 15-45 | Medium-Leptin |
| | 40-70 | High-Leptin |
| | 75-100 | Very High Scaling |

### a. MEMBERSHIP FUNCTIONS OF LEPTIN

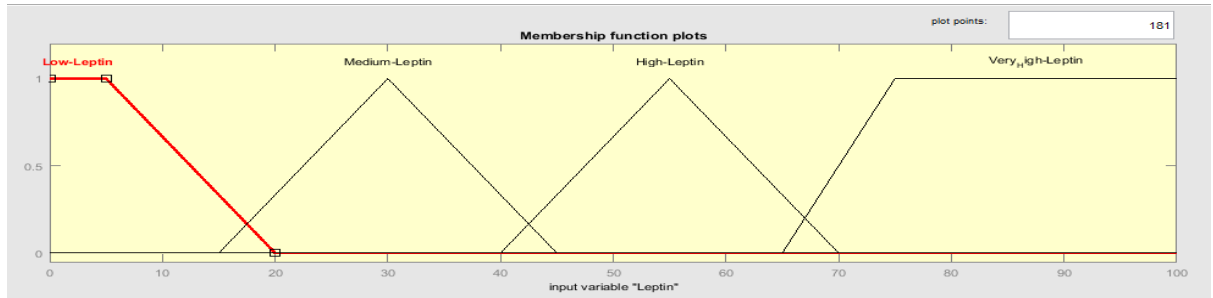

Figure 5: Membership functions of Leptin

### b. EQUATIONS (4) OF MEMBERSHIP EXPRESSIONS FOR LEPTIN (O,D)

$$\mu_{\text{Low-Leptin}}(x) = \begin{cases} 1 & x < 5 \\ \frac{20-x}{15} & 5 \leq x < 20 \end{cases}$$

$$\mu_{\text{Medium-Leptin}}(x) = \begin{cases} \frac{x-15}{15} & 15 \leq x < 30 \\ 1 & x = 30 \\ \frac{45-x}{15} & 30 \leq x < 45 \end{cases}$$

$$\mu_{\text{High-Leptin}}(x) = \begin{cases} \frac{x-40}{15} & 40 \leq x < 55 \\ 1 & x = 55 \\ \frac{70-x}{15} & 55 \leq x < 70 \end{cases}$$

$$\mu_{\text{Very High-Leptin}}(x) = \begin{cases} \frac{x-65}{15} & 65 \leq x < 75 \\ 1 & x \geq 75 \end{cases}$$

### c. FUZZY SOFT SET OF INSULIN (Õ,D)

Table 9 illustrates a fuzzy soft set for the Insulin variable.

*Table 8: Fuzzy Soft Set of Leptin (Õ,D)*

| Sample No | (LPN)$_L$=$\delta_1$ | (LPN)$_M$=$\delta_2$ | (LPN)$_H$=$\delta_3$ | (LPN)$_{VH}$=$\delta_4$ |
|---|---|---|---|---|
| $\mu_3$ | 0.17 | 0.16 | 0.00 | 0.00 |
| $\mu_{11}$ | 0.00 | 0.62 | 0.00 | 0.00 |
| $\mu_{19}$ | 0.00 | 0.41 | 0.00 | 0.00 |
| $\mu_{31}$ | 0.00 | 0.00 | 0.64 | 0.00 |
| $\mu_{45}$ | 0.00 | 0.00 | 0.99 | 0.00 |
| $\mu_{60}$ | 0.68 | 0.00 | 0.00 | 0.00 |
| $\mu_{71}$ | 0.00 | 0.21 | 0.00 | 0.00 |
| $\mu_{82}$ | 0.06 | 0.27 | 0.00 | 0.00 |
| $\mu_{91}$ | 0.00 | 0.89 | 0.00 | 0.00 |
| $\mu_{104}$ | 0.00 | 0.78 | 0.00 | 0.00 |

### E. ADIPONECTIN

Reduced adiponectin levels have been implicated in the development of breast cancer in obese patients. We split adiponectin into three fuzzy sets, as shown in table 10, namely Low-Adiponectin, Medium-Adiponectin, and High-Adiponectin. Membership functions and membership expressions are depicted in Fig 6, as well as in Equation (5). The fuzzy soft sets for Adiponectin are shown in Table 11.

*Table 9: Classification of Adiponectin*

| Symptom | Range | Fuzzy Set |
|---|---|---|
| ***Adiponectin*** | 0-10 | Low-Adiponectin |
|  | 7-23 | Medium-Adiponectin |
|  | 20-40 | High-Adiponectin |

### a. MEMBERSHIP FUNCTIONS OF ADIPONECTIN

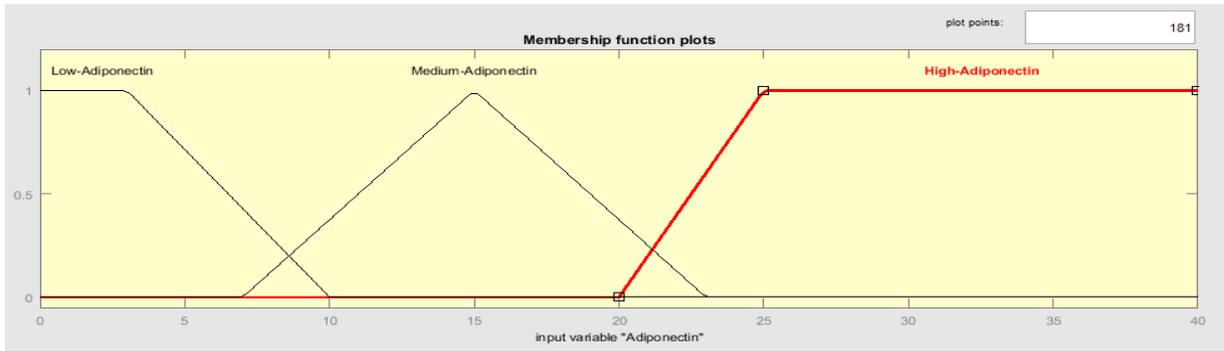

*Figure 6: Membership Functions of Adiponectin*

### b. FUZZY MEMBERSHIP EXPRESSIONS (5) FOR ADIPONECTIN

$$\mu_{\text{Low-Adiponectin}}(x) = \begin{cases} 1 & x < 3 \\ \frac{10-x}{7} & 3 \leq x < 10 \end{cases}$$

$$\mu_{\text{Medium-Adiponectin}}(x) = \begin{cases} \frac{x-7}{8} & 7 \leq x < 15 \\ 1 & x = 15 \\ \frac{23-x}{8} & 15 \leq x < 23 \end{cases}$$

$$\mu_{\text{High-Adiponectin}}(x) = \begin{cases} \frac{x-20}{5} & 20 \leq x < 25 \\ 1 & x \geq 25 \end{cases}$$

### c. FUZZY SOFT SET OF ADIPONECTIN $(\tilde{P},E)$

Table 9 illustrates a fuzzy soft set for the Adiponectin variable.

*Table 10: Fuzzy Soft Set of Adiponectin $(\tilde{P},E)$*

| Sample No | $(ADP)_L = \varepsilon_1$ | $(ADP)_M = \varepsilon_2$ | $(ADP)_H = \varepsilon_3$ |
|---|---|---|---|
| $\mu_3$ | 0.00 | 0.07 | 0.48 |
| $\mu_{11}$ | 0.00 | 0.00 | 1.00 |
| $\mu_{19}$ | 0.64 | 0.00 | 0.00 |
| $\mu_{31}$ | 0.35 | 0.06 | 0.00 |
| $\mu_{45}$ | 0.26 | 0.14 | 0.00 |
| $\mu_{60}$ | 0.00 | 0.53 | 0.00 |
| $\mu_{71}$ | 0.00 | 0.86 | 0.00 |
| $\mu_{82}$ | 0.65 | 0.00 | 0.00 |
| $\mu_{91}$ | 0.02 | 0.36 | 0.00 |
| $\mu_{104}$ | 1.00 | 0.00 | 0.00 |

### F. RESULTANT SOFT SET OF AGE AND BODY MASS INDEX

Table 12 illustrates the result of the MAX (AND) operation.

*Table 11: Resultant Soft Set of Age and Body Mass Index*

| Sample No | 1,1 | 1,2 | 1,3 | 2,1 | 2,2 | 2,3 | 3,1 | 3,2 | 3,3 | 4,1 | 4,2 | 4,3 |
|---|---|---|---|---|---|---|---|---|---|---|---|---|
| $\mu_3$ | 0.00 | 0.50 | 0.00 | 0.00 | 0.50 | 0.00 | 0.00 | 0.50 | 0.00 | 1.00 | 1.00 | 1.00 |
| $\mu_{11}$ | 0.00 | 0.50 | 0.00 | 0.00 | 0.50 | 0.00 | 0.73 | 0.73 | 0.73 | 0.00 | 0.50 | 0.00 |
| $\mu_{19}$ | 0.00 | 0.00 | 0.80 | 0.00 | 0.00 | 0.80 | 0.00 | 0.00 | 0.80 | 0.93 | 0.93 | 0.93 |
| $\mu_{31}$ | 0.00 | 0.00 | 1.00 | 0.00 | 0.00 | 1.00 | 0.00 | 0.00 | 1.00 | 1.00 | 1.00 | 1.00 |
| $\mu_{45}$ | 0.00 | 0.83 | 0.00 | 0.00 | 0.83 | 0.00 | 0.00 | 0.83 | 0.00 | 1.00 | 1.00 | 1.00 |
| $\mu_{60}$ | 0.00 | 0.33 | 0.00 | 0.00 | 0.33 | 0.00 | 0.00 | 0.33 | 0.00 | 0.80 | 0.80 | 0.80 |

| | | | | | | | | | | | |
|---|---|---|---|---|---|---|---|---|---|---|---|
| μ71 | 0.00 | 0.80 | 0.00 | 0.00 | 0.80 | 0.00 | 0.93 | 0.93 | 0.93 | 0.00 | 0.80 | 0.00 |
| μ82 | 0.00 | 0.83 | 0.00 | 0.00 | 0.83 | 0.00 | 0.00 | 0.83 | 0.00 | 1.00 | 1.00 | 1.00 |
| μ91 | 0.00 | 0.35 | 0.24 | 0.00 | 0.35 | 0.24 | 0.00 | 0.35 | 0.24 | 1.00 | 1.00 | 1.00 |
| μ104 | 0.00 | 0.00 | 0.96 | 0.00 | 0.00 | 0.96 | 0.20 | 0.20 | 0.96 | 0.46 | 0.46 | 0.96 |

The membership values of factor ten (10) patients.

*Table 12: The Membership Value of Ten Patients*

| Sample No | Age | BMI | Insulin | Leptin | Adiponectin |
|---|---|---|---|---|---|
| μ3 | 1.0 O | 0.5 OII | 0.1 L 0.42 N | 0.17 L 0.16 M | 0.07M 0.48H |
| μ11 | 0.73 M | 0.5 OII | 0.76 N | 0.62 M | 1.0 H |
| μ19 | 0.93 O | 0.8 OIII | 0.11 L 0.40 N | 0.41 M | 0.64 L |
| μ31 | 1.0 O | 1.0 OIII | 1.0 H | 0.64 H | 0.35 L 0.06 M |
| μ45 | 1.0 O | 0.83 OII | 0.47 N 0.14H | 0.9 H | 0.26 L 0.14 M |
| μ60 | 0.8O | 0.33 OII | 0.43 L 0.13 N | 0.68 L | 0.53 m |
| μ71 | 0.93 M | 0.8 OII | 1.0 H | 0.21 M | 0.86 M |
| μ82 | 1.0 O | 0.83 OII | 1.0 H | 0.06 L 0.27M | 0.65 L |
| μ91 | 1.0 O | 0.35 OII 0.24 OIII | 1.0 H | 0.89 M | 0.02 L 0.36 M |
| μ104 | 0.2 M 0.46 O | 0.96 OIII | 1.0 H | 0.78 M | 1.0 L |

## 4. Second step: transforming the fuzzy sets to soft sets

We already know that any fuzzy set can be thought of as a soft set. To begin, we use the membership functions to select the parameter set. As a result, we have numerical values for each parameter in a parameter set. Several of the soft sets obtained through the fuzzy set relation. A fuzzy soft set is the result of combining fuzzy set and soft set theory. As a result, we can convert a fuzzy set to a fuzzy soft set. Let U={μ3, μ11, $\mu25$, μ40, μ50, μ59, μ71, μ77, μ91, μ104} be the set of ten patients and the parameter set, E ={(AGE)$_C$, (AGE)$_Y$, (AGE)$_M$, (AGE)$_O$, (BMI)$_{OI}$, (BMI)$_{OII}$, (BMI)$_{OIII}$, (INS)$_L$, (INS)$_M$, (INS)$_H$, (LPN)$_L$, (LPN)$_M$, (LPN)$_H$, (LPN)$_{VH}$, (ADP)$_L$, (ADP)$_M$, (ADP)$_H$}. Let A, B, C and D, denote four subsets of the set of parameters E, where A={(AGE)$_C$, (AGE)$_Y$, (AGE)$_M$, (AGE)$_O$}, B={(BMI)$_{OI}$, (BMI)$_{OII}$, (BMI)$_{OIII}$}, C= {(INS)$_L$, (INS)$_M$, (INS)$_H$}, D={(LPN)$_L$, (LPN)$_M$, (LPN)$_H$, (LPN)$_{VH}$} and D={(ADP)$_L$, (ADP)$_M$, (ADP)$_H$} Assume that the fuzzy soft (L̃, A) describes the 'Age', the fuzzy soft set (M̃,B) describes the 'BMI', the fuzzy soft set (Ñ,C) describes the 'INS' the fuzzy soft (Õ,D) describes the 'LPN' and the fuzzy soft set (P̃,E) describes the 'ADP '. From Table 13, we obtain the fuzzy soft set as follows:

(L̃, A) ={(AGE)$_C$ = {μ3/.0, μ11/.0, $\mu25$/.0, μ40/.0, μ50/.0, μ59/.0, μ71/.0, μ77/.0, μ91/.0, μ104/.0},
(AGE)$_Y$ = {μ3/0.0, μ11/0.0, $\mu25$/0.0, μ40/0.0, μ50/0.0, μ59/0.0, μ71/0.0, μ77/0.0, μ91/0.0, μ104/0.0},
(AGE)$_M$ = {μ3/0.0, μ11/0.73, $\mu25$/1.0, μ40/0.0, μ50/0.0, μ59/0.0, μ71/0.93, μ77/0.86, μ91/0.0, μ104/0.20},
(AGE)$_O$ = { μ3/1.0, μ11/0.0, $\mu25$/0.0, μ40/1.0, μ50/1.0, μ59/0.80, μ71/0.0, μ77/0.0, μ91/1.0, μ104/0.46}}

(M̃,B) = {(BMI)$_{OI}$ = { μ3/0.48, μ11/1.0, $\mu25$/0.0, μ40/0.0, μ50/0.0, μ59/0.0, μ71/0.0, μ77/0.0, μ91/0.0, μ104/0.0},
(BMI)$_{OII}$ = {μ3/0.50, μ11/0.50, $\mu25$/0.00, μ40/0.57, μ50/0.85, μ59/0.30, μ71/0.80, μ77/0.25, μ91/0.35, μ104/0.0},
 (BMI)$_{OIII}$ = {μ3/0.0, μ11/0.0, $\mu25$/1.0, μ40/0.0, μ50/0.0, μ59/0.0, μ71/0.0, μ77/0.25, μ91/0.24, μ104/0.96}}

(Ñ,C) = {(INS)$_L$={ μ3/0.10, μ11/0.0, $\mu25$/0.0, μ40/0.0, μ50/1.0, μ59/0.43, μ71/0.0, μ77/0.0, μ91/0.0, μ104/0.0},
(INS)$_M$ = {μ3/0.42, μ11/0.76, $\mu25$/0.92, μ40/0.68, μ50/0.0, μ59/0.13, μ71/0.0, μ77/0.44, μ91/0.0, μ104/0.0},
(INS)$_H$ = {μ3/0.0, μ11/0.0, $\mu25$/0.0, μ40/0.0, μ50/0.0, μ59/0.0, μ71/1.0, μ77/0.0, μ91/1.0, μ104/1.0}}

(Õ,D) = {(LPN)$_L$={ μ3/0.17, μ11/0.0, $\mu25$/0.0, μ40/0.0, μ50/0.0, μ59/0.67, μ71/0.0, μ77/0.0, μ91/0.0, μ104/0.0},
(LPN)$_M$ = {μ3/0.16, μ11/0.62, $\mu25$/0.33, μ40/0.90, μ50/0.45, μ59/0.0, μ71/0.21, μ77/0.71, μ91/0.89, μ104/0.78},
(LPN)$_H$ = {μ3/0.0, μ11/0.0, $\mu25$/0.0, μ40/0.0, μ50/0.0, μ59/0.0, μ71/0.0, μ77/0.0, μ91/0.0, μ104/0.0},
(LPN)$_{VH}$ = {μ3/0.0, μ11/0.0, $\mu25$/0.0, μ40/0.0, μ50/0.0, μ59/0.0, μ71/0.0, μ77/0.0, μ91/0.0, μ104/0.0}}

(P̃,E) = {(ADP)$_L$={μ3/0.0, μ11/0.0, $\mu25$/0.76, μ40/0.37, μ50/0.72, μ59/0.0, μ71/0.0, μ77/0.0, μ91/0.02, μ104/1.0},
(ADP)$_M$ = {μ3/0.07, μ11/0.0, $\mu25$/0.0, μ40/0.0, μ50/0.0, μ59/0.53, μ71/0.86, μ77/0.71, μ91/0.36, μ104/0.0},
(ADP)$_H$ = {μ3/0.48, μ11/1.0, $\mu25$/0.0, μ40/0.0, μ50/0.0, μ59/0.0, μ71/0.0, μ77/0.0, μ91/0.0, μ104/0.0}}

## 5. Normal Parameter Reduction of Fuzzy Soft Sets

Parameter lowering is critical when confronted with a decision-making problem [17]. This process efficiently reduces the number of parameters in a problem, emphasizing only the critical parameters.

Suppose $U=\{h_1, h_2,…,h_n\}$, $E=\{e_1,e_2,…,e_m\}$ and $(F,E)$ is a soft set with tabular representation. Let $fE(h_i)=\sum_j h_{ij}$ where $h_{ij}$ are the entries in the table of $(F,E)$. Further we use ME to denote the collection of objects in U which takes the max value of fE. For every $A \subset E$, if $ME−A=ME$, then A is called a dispensable set in E, otherwise A is called an indispensable set in E. Roughly speaking, $A \subset E$ is dispensable means that the difference among all objects according to the parameters in A does not influence the final decision. The parameter set E is called independent if every proper subset of E is indispensable, otherwise E is dependent. $B \subseteq E$ is called a reduction of E if B is independent and $MB=ME$, i.e., B is the minimal subset of E that keeps the optimal choice objects invariant.

$(\tilde{L}, A)$ = {(AGE)$_M$= {µ3/0.0, µ11/0.0, $\mu25$/1.0, $\mu40$/0.0, µ50/0.0, µ59/0.0, µ71/0.93, µ77/0.86, µ91/0.0, µ104/0.20},
(AGE)$_O$={µ3/1.0, µ11/0.0, $\mu25$/0.0, $\mu40$/1.0, µ50/1.0, µ59/0.80, µ71/0.0, µ77/0.0, µ91/1.0, µ104/0.46}}

$(\tilde{M},B)$= {(BMI)$_{OI}$={ µ3/0.48, µ11/1.0, $\mu25$/0.0, $\mu40$/0.0, µ50/0.0, µ59/0.0, µ71/0.0, µ77/0.0, µ91/0.0, µ104/0.0},
(BMI)$_{OII}$ ={µ3/0.50, µ11/0.50, $\mu25$/0.00, $\mu40$/0.57, µ50/0.85, µ59/0.30, µ71/0.80, µ77/0.25, µ91/0.35, µ104/0.0},
(BMI)$_{OIII}$={ µ3/0.0, µ11/0.0, $\mu25$/1.0, $\mu40$/0.0, µ50/0.0, µ59/0.0, µ71/0.0, µ77/0.25, µ91/0.24, µ104/0.96}}

$(\tilde{N},C)$ = { (INS)$_L$={ µ3/0.10, µ11/0.0, $\mu25$/0.0, $\mu40$/0.0, µ50/1.0, µ59/0.43, µ71/0.0, µ77/0.0, µ91/0.0, µ104/0.0},
(INS)$_M$ = {µ3/0.42, µ11/0.76, $\mu25$/0.92, $\mu40$/0.68, µ50/0.0, µ59/0.13, µ71/0.0, µ77/0.44, µ91/0.0, µ104/0.0},
(INS)$_H$ = {µ3/0.0, µ11/0.0, $\mu25$/0.0, $\mu40$/0.0, µ50/0.0, µ59/0.0, µ71/1.0, µ77/0.0, µ91/1.0, µ104/1.0}}

$(\tilde{O},D)$ = { (LPN)$_L$={µ3/0.17, µ11/0.0, $\mu25$/0.0, $\mu40$/0.0, µ50/0.0, µ59/0.67, µ71/0.0, µ77/0.0, µ91/0.0, µ104/0.0},
(LPN)$_M$ = {µ3/0.16, µ11/0.62, $\mu25$/0.33, $\mu40$/0.90, µ50/0.45, µ59/0.0, µ71/0.21, µ77/0.71, µ91/0.89, µ104/0.78}}

$(\tilde{P},E)$ = {(ADP)$_L$={µ3/0.0, µ11/0.0, $\mu25$/0.76, $\mu40$/0.37, µ50/0.72, µ59/0.0, µ71/0.0, µ77/0.0, µ91/0.02, µ104/1.0},
(ADP)$_M$ = {µ3/0.07, µ11/0.0, $\mu25$/0.0, $\mu40$/0.0, µ50/0.0, µ59/0.53, µ71/0.86, µ77/0.71, µ91/0.36, µ104/0.0},
(ADP)$_H$ = {µ3/0.48, µ11/1.0, $\mu25$/0.0, $\mu40$/0.0, µ50/0.0, µ59/0.0, µ71/0.0, µ77/0.0, µ91/0.0, µ104/0.0}}

## 6. ALGORITHM

By utilizing Kong's algorithm, we can forecast which patients will develop breast cancer [18].

| Sample No | €1 | €2 | €3 | €4 | €5 | €6 | €7 | €8 | €9 | €10 | €11 | €12 | €13 | €14 | €15 |
|---|---|---|---|---|---|---|---|---|---|---|---|---|---|---|---|
| $µ_3$ | 0.50 | 0.50 | 0.50 | 0.50 | 0.50 | 0.50 | 0.50 | 0.50 | 0.50 | 0.50 | 0.50 | 0.50 | 0.50 | 0.50 | 0.50 |
| $µ_{11}$ | 0.73 | 0.73 | 1.00 | 0.73 | 0.73 | 1.00 | 0.76 | 0.76 | 1.00 | 0.76 | 0.76 | 1.00 | 0.73 | 0.73 | 1.00 |
| $µ_{19}$ | 0.64 | 0.11 | 0.11 | 0.64 | 0.41 | 0.41 | 0.64 | 0.40 | 0.40 | 0.64 | 0.41 | 0.41 | 0.64 | 0.00 | 0.00 |
| $µ_{31}$ | 0.35 | 0.06 | 0.00 | 0.64 | 0.64 | 0.64 | 0.35 | 0.06 | 0.00 | 0.64 | 0.64 | 0.64 | 1.00 | 1.00 | 1.00 |
| $µ_{45}$ | 0.83 | 0.83 | 0.83 | 0.99 | 0.99 | 0.99 | 0.83 | 0.83 | 0.83 | 0.99 | 0.99 | 0.99 | 0.83 | 0.83 | 0.83 |
| $µ_{60}$ | 0.68 | 0.68 | 0.68 | 0.43 | 0.53 | 0.43 | 0.68 | 0.68 | 0.68 | 0.33 | 0.53 | 0.33 | 0.68 | 0.68 | 0.68 |
| $µ_{71}$ | 0.93 | 0.93 | 0.93 | 0.93 | 0.93 | 0.93 | 0.93 | 0.93 | 0.93 | 0.93 | 0.93 | 0.93 | 1.00 | 1.00 | 1.00 |
| $µ_{82}$ | 0.83 | 0.83 | 0.83 | 0.83 | 0.83 | 0.83 | 0.83 | 0.83 | 0.83 | 0.83 | 0.83 | 0.83 | 1.00 | 1.00 | 1.00 |
| $µ_{91}$ | 0.35 | 0.36 | 0.35 | 0.89 | 0.89 | 0.89 | 0.35 | 0.36 | 0.35 | 0.89 | 0.89 | 0.89 | 1.00 | 1.00 | 1.00 |
| $µ_{104}$ | 1.00 | 0.20 | 0.20 | 1.00 | 0.78 | 0.78 | 1.00 | 0.20 | 0.20 | 1.00 | 0.78 | 0.78 | 1.00 | 1.00 | 1.00 |

| Sample No | €16 | €17 | €18 | €19 | €20 | €21 | €22 | €23 | €24 | €25 | €26 | €27 | €28 | €29 | €30 |
|---|---|---|---|---|---|---|---|---|---|---|---|---|---|---|---|
| $µ_3$ | 0.50 | 0.50 | 0.50 | 0.17 | 0.17 | 0.48 | 0.16 | 0.16 | 0.48 | 0.42 | 0.42 | 0.48 | 0.42 | 0.42 | 0.48 |
| $µ_{11}$ | 0.73 | 0.73 | 1.00 | 0.73 | 0.73 | 1.00 | 0.73 | 0.73 | 1.00 | 0.76 | 0.76 | 1.00 | 0.76 | 0.76 | 1.00 |
| $µ_{19}$ | 0.64 | 0.41 | 0.41 | 0.80 | 0.80 | 0.80 | 0.80 | 0.80 | 0.80 | 0.80 | 0.80 | 0.80 | 0.80 | 0.80 | 0.80 |
| $µ_{31}$ | 1.00 | 1.00 | 1.00 | 1.00 | 1.00 | 1.00 | 1.00 | 1.00 | 1.00 | 1.00 | 1.00 | 1.00 | 1.00 | 1.00 | 1.00 |
| $µ_{45}$ | 0.99 | 0.99 | 0.99 | 0.26 | 0.14 | 0.00 | 0.99 | 0.99 | 0.99 | 0.47 | 0.47 | 0.47 | 0.99 | 0.99 | 0.99 |
| $µ_{60}$ | 0.33 | 0.53 | 0.33 | 0.68 | 0.68 | 0.68 | 0.43 | 0.53 | 0.43 | 0.68 | 0.68 | 0.68 | 0.13 | 0.53 | 0.13 |

| | | | | | | | | | | | | | | |
|---|---|---|---|---|---|---|---|---|---|---|---|---|---|---|
| $\mu_{71}$ | 1.00 | 1.00 | 1.00 | 0.93 | 0.93 | 0.93 | 0.93 | 0.93 | 0.93 | 0.93 | 0.93 | 0.93 | 0.93 | 0.93 | 0.93 |
| $\mu_{82}$ | 1.00 | 1.00 | 1.00 | 0.65 | 0.06 | 0.06 | 0.65 | 0.27 | 0.27 | 0.65 | 0.06 | 0.06 | 0.65 | 0.27 | 0.27 |
| $\mu_{91}$ | 1.00 | 1.00 | 1.00 | 0.24 | 0.36 | 0.24 | 0.89 | 0.89 | 0.89 | 0.24 | 0.36 | 0.24 | 0.89 | 0.89 | 0.89 |
| $\mu_{104}$ | 1.00 | 1.00 | 1.00 | 1.00 | 0.96 | 0.96 | 1.00 | 0.96 | 0.96 | 1.00 | 0.96 | 0.96 | 1.00 | 0.96 | 0.96 |

| Sample No | €31 | €32 | €33 | €34 | €35 | €36 | €37 | €38 | €39 | €40 | €41 | €42 | €43 | €44 | €45 |
|---|---|---|---|---|---|---|---|---|---|---|---|---|---|---|---|
| $\mu_3$ | 0.17 | 0.17 | 0.48 | 0.16 | 0.16 | 0.48 | 1.00 | 1.00 | 1.00 | 1.00 | 1.00 | 1.00 | 1.00 | 1.00 | 1.00 |
| $\mu_{11}$ | 0.73 | 0.73 | 1.00 | 0.73 | 0.73 | 1.00 | 0.50 | 0.50 | 1.00 | 0.62 | 0.62 | 1.00 | 0.76 | 0.76 | 1.00 |
| $\mu_{19}$ | 0.80 | 0.80 | 0.80 | 0.80 | 0.80 | 0.80 | 0.93 | 0.93 | 0.93 | 0.93 | 0.93 | 0.93 | 0.93 | 0.93 | 0.93 |
| $\mu_{31}$ | 1.00 | 1.00 | 1.00 | 1.00 | 1.00 | 1.00 | 1.00 | 1.00 | 1.00 | 1.00 | 1.00 | 1.00 | 1.00 | 1.00 | 1.00 |
| $\mu_{45}$ | 0.26 | 0.14 | 0.14 | 0.99 | 0.99 | 0.99 | 1.00 | 1.00 | 1.00 | 1.00 | 1.00 | 1.00 | 1.00 | 1.00 | 1.00 |
| $\mu_{60}$ | 0.68 | 0.68 | 0.68 | 0.00 | 0.53 | 0.00 | 0.80 | 0.80 | 0.80 | 0.80 | 0.80 | 0.80 | 0.80 | 0.80 | 0.80 |
| $\mu_{71}$ | 1.00 | 1.00 | 1.00 | 1.00 | 1.00 | 1.00 | 0.80 | 0.86 | 0.80 | 0.80 | 0.86 | 0.80 | 0.80 | 0.86 | 0.80 |
| $\mu_{82}$ | 1.00 | 1.00 | 1.00 | 1.00 | 1.00 | 1.00 | 1.00 | 1.00 | 1.00 | 1.00 | 1.00 | 1.00 | 1.00 | 1.00 | 1.00 |
| $\mu_{91}$ | 1.00 | 1.00 | 1.00 | 1.00 | 1.00 | 1.00 | 1.00 | 1.00 | 1.00 | 1.00 | 1.00 | 1.00 | 1.00 | 1.00 | 1.00 |
| $\mu_{104}$ | 1.00 | 1.00 | 1.00 | 1.00 | 1.00 | 1.00 | 1.00 | 0.46 | 0.46 | 1.00 | 0.78 | 0.78 | 1.00 | 0.46 | 0.46 |

| Sample No | €46 | €47 | €48 | €49 | €50 | €51 | €52 | €53 | €54 | €55 | €56 | €57 | €58 | €59 | €60 |
|---|---|---|---|---|---|---|---|---|---|---|---|---|---|---|---|
| $\mu_3$ | 1.00 | 1.00 | 1.00 | 1.00 | 1.00 | 1.00 | 1.00 | 1.00 | 1.00 | 1.00 | 1.00 | 1.00 | 1.00 | 1.00 | 1.00 |
| $\mu_{11}$ | 0.76 | 0.76 | 1.00 | 0.50 | 0.50 | 1.00 | 0.62 | 0.62 | 1.00 | 0.00 | 0.00 | 1.00 | 0.62 | 0.62 | 1.00 |
| $\mu_{19}$ | 0.93 | 0.93 | 0.93 | 0.93 | 0.93 | 0.93 | 0.93 | 0.93 | 0.93 | 0.93 | 0.93 | 0.93 | 0.93 | 0.93 | 0.93 |
| $\mu_{31}$ | 1.00 | 1.00 | 1.00 | 1.00 | 1.00 | 1.00 | 1.00 | 1.00 | 1.00 | 1.00 | 1.00 | 1.00 | 1.00 | 1.00 | 1.00 |
| $\mu_{45}$ | 1.00 | 1.00 | 1.00 | 1.00 | 1.00 | 1.00 | 1.00 | 1.00 | 1.00 | 1.00 | 1.00 | 1.00 | 1.00 | 1.00 | 1.00 |
| $\mu_{60}$ | 0.80 | 0.80 | 0.80 | 0.80 | 0.80 | 0.80 | 0.80 | 0.80 | 0.80 | 0.80 | 0.80 | 0.80 | 0.80 | 0.80 | 0.80 |
| $\mu_{71}$ | 0.80 | 0.86 | 0.80 | 1.00 | 1.00 | 1.00 | 1.00 | 1.00 | 1.00 | 0.00 | 0.86 | 0.00 | 0.21 | 0.86 | 0.21 |
| $\mu_{82}$ | 1.00 | 1.00 | 1.00 | 1.00 | 1.00 | 1.00 | 1.00 | 1.00 | 1.00 | 1.00 | 1.00 | 1.00 | 1.00 | 1.00 | 1.00 |
| $\mu_{91}$ | 1.00 | 1.00 | 1.00 | 1.00 | 1.00 | 1.00 | 1.00 | 1.00 | 1.00 | 1.00 | 1.00 | 1.00 | 1.00 | 1.00 | 1.00 |
| $\mu_{104}$ | 1.00 | 0.78 | 0.78 | 1.00 | 1.00 | 1.00 | 1.00 | 1.00 | 1.00 | 1.00 | 0.96 | 0.96 | 1.00 | 0.96 | 0.96 |

| Sample No | €61 | €62 | €63 | €64 | €65 | €66 | €67 | €68 | €69 | €70 | €71 | €72 |
|---|---|---|---|---|---|---|---|---|---|---|---|---|
| $\mu_3$ | 1.00 | 1.00 | 1.00 | 1.00 | 1.00 | 1.00 | 1.00 | 1.00 | 1.00 | 1.00 | 1.00 | 1.00 |
| $\mu_{11}$ | 0.76 | 0.76 | 1.00 | 0.76 | 0.76 | 1.00 | 0.00 | 0.00 | 1.00 | 0.62 | 0.62 | 1.00 |
| $\mu_{19}$ | 0.93 | 0.93 | 0.93 | 0.93 | 0.93 | 0.93 | 0.93 | 0.93 | 0.93 | 0.93 | 0.93 | 0.93 |
| $\mu_{31}$ | 1.00 | 1.00 | 1.00 | 1.00 | 1.00 | 1.00 | 1.00 | 1.00 | 1.00 | 1.00 | 1.00 | 1.00 |
| $\mu_{45}$ | 1.00 | 1.00 | 1.00 | 1.00 | 1.00 | 1.00 | 1.00 | 1.00 | 1.00 | 1.00 | 1.00 | 1.00 |
| $\mu_{60}$ | 0.80 | 0.80 | 0.80 | 0.80 | 0.80 | 0.80 | 0.80 | 0.80 | 0.80 | 0.80 | 0.80 | 0.80 |
| $\mu_{71}$ | 0.00 | 0.86 | 0.00 | 0.21 | 0.86 | 0.21 | 1.00 | 1.00 | 1.00 | 1.00 | 1.00 | 1.00 |
| $\mu_{82}$ | 1.00 | 1.00 | 1.00 | 1.00 | 1.00 | 1.00 | 1.00 | 1.00 | 1.00 | 1.00 | 1.00 | 1.00 |
| $\mu_{91}$ | 1.00 | 1.00 | 1.00 | 1.00 | 1.00 | 1.00 | 1.00 | 1.00 | 1.00 | 1.00 | 1.00 | 1.00 |
| $\mu_{104}$ | 1.00 | 0.96 | 0.96 | 1.00 | 0.96 | 0.96 | 1.00 | 1.00 | 1.00 | 1.00 | 1.00 | 1.00 |

## 7. COMPARISON TABLE

| Sample No | $\mu_3$ | $\mu_{11}$ | $\mu_{19}$ | $\mu_{31}$ | $\mu_{45}$ | $\mu_{60}$ | $\mu_{71}$ | $\mu_{82}$ | $\mu_{91}$ | $\mu_{104}$ |
|---|---|---|---|---|---|---|---|---|---|---|
| $\mu_3$ | 72 | 36 | 48 | 48 | 48 | 42 | 36 | 43 | 46 | 40 |
| $\mu_{11}$ | 48 | 72 | 36 | 32 | 30 | 64 | 28 | 32 | 32 | 28 |
| $\mu_{19}$ | 24 | 36 | 72 | 12 | 12 | 63 | 24 | 12 | 10 | 12 |
| $\mu_{31}$ | 60 | 60 | 60 | 72 | 60 | 60 | 60 | 60 | 62 | 60 |
| $\mu_{45}$ | 48 | 54 | 60 | 48 | 72 | 62 | 36 | 52 | 48 | 44 |
| $\mu_{60}$ | 30 | 8 | 9 | 12 | 11 | 72 | 8 | 6 | 12 | 8 |
| $\mu_{71}$ | 48 | 53 | 48 | 36 | 48 | 64 | 72 | 48 | 48 | 40 |
| $\mu_{82}$ | 65 | 56 | 60 | 60 | 64 | 66 | 48 | 72 | 56 | 56 |
| $\mu_{91}$ | 62 | 56 | 62 | 60 | 64 | 60 | 48 | 64 | 72 | 56 |
| $\mu_{104}$ | 52 | 52 | 62 | 48 | 48 | 64 | 56 | 48 | 48 | 72 |

## 8. SCORE TABLE

| Sample No | Row Sum | Column Sum | Score |
|---|---|---|---|
| $\mu_3$ | 459 | 509 | -50 |
| $\mu_{11}$ | 402 | 483 | -81 |
| $\mu_{19}$ | 277 | 517 | -240 |
| $\mu_{31}$ | 614 | 428 | 186 |
| $\mu_{45}$ | 524 | 457 | 67 |
| $\mu_{60}$ | 176 | 617 | -441 |
| $\mu_{71}$ | 505 | 416 | 89 |
| $\mu_{82}$ | 603 | 437 | 166 |
| $\mu_{91}$ | 604 | 434 | 170 |
| $\mu_{104}$ | 550 | 416 | 134 |

## 9. RESULTS AND DISCUSSION

Breast cancer is one of the most common types of cancer in women globally. The detection of these cancers in patients using soft computing techniques may enable the automatic breast cancer diagnostic regime to be expanded. False-positive cases are common in fibrous breast tissues; locating masses using a variety of radiological modalities is much more difficult than locating calcifications due to their significantly different sizes and shapes [19]. Machine learning assessment and decision-making for clinical and radiological diagnosis are critical [20-22]. Soft computing and fuzzy set theory may assist physicians in identifying risk that is not visible using conventional methods. Early detection of breast cancer has been shown to save lives and assist physicians in initiating treatment early.

The fuzzy soft set theory exemplifies the physiognomies of rapid and accurate learning with the capacity to use both linguistic and non-linguistic data and an exceptional capacity for generalization. Soft expert systems are a rule-based method for rapidly constructing previously extracted connections and record-based measures to assist physicians or surgeons in making timely decisions.

To develop a soft expert system capable of serving as a predictive model for breast cancer the factors of patients' age, body mass index, insulin level, leptin level, and adiponectin level are used to assess risk using fuzzy rules and soft set theory.

A higher score in the score table indicates a greater risk of breast cancer. From dataset the patients $\mu_3$, $\mu_{11}$, $\mu_{19}$, $\mu_{31}$ and $\mu_{45}$ are classified as healthy control and patients $\mu_{60}$, $\mu_{71}$, $\mu_{82}$, $\mu_{91}$ and $\mu_{104}$ are patients with high risk. If accurate, prediction models based on predictors may be used as a biomarker for breast cancer. Our model predicts $\mu_3$, $\mu_{11}$, $\mu_{19}$, $\mu_{71}$, $\mu_{82}$, $\mu_{91}$, $\mu_{104}$ (7 Patients) correctly, and, $\mu_{31}$. $\mu_{45}$ and $\mu_{60}$ wrongly. Hence the accuracy of this model [8] is 70%.

Further research should be conducted to extend this method by utilizing the Fuzzy Rough Nearest Neighbors classification and prediction paradigm to deal with inherent uncertainty in the dataset [23]. The nearest neighbor (NN) algorithm is inspired by how humans make decisions based on comparisons between test objects and previously encountered samples. This technique can be used in conjunction with a neural network algorithm to classify or predict the decision value of test objects by utilizing lower and upper approximations from fuzzy-rough set theory.

**Author Contributions:** All authors listed have made a substantial, direct, and intellectual contribution to the work and approved it for publication.

**Funding:** "This research received no external funding."

**Conflicts of Interest:** "The authors declare no conflict of interest."